\documentclass[preprint,12pt]{elsarticle}

\usepackage{amssymb}
\usepackage[T1]{fontenc}
\usepackage{xspace}
\usepackage{subcaption}
\usepackage{listings}
\usepackage{comment}
\usepackage[ruled, linesnumbered]{algorithm2e}
\usepackage{soul,color}
\def\ourmethod{\textsc{iEvoFlow}\xspace}

\def\ie{i.e.,\xspace}
\def\eg{e.g.,\xspace}

\journal{anonymized}

\begin{document}

\begin{frontmatter}

\title{Evolving machine learning workflows through interactive AutoML}

\author[uco,dasci]{Rafael Barbudo}
\author[uma]{Aurora Ramírez}
\author[uco,dasci]{José Raúl Romero\corref{cor1}}

\affiliation[uco]{
            organization={Department of Computer Science and Numerical Analysis, University of C\'ordoba},
            city={C\'ordoba},
            postcode={14071}, 
            country={Spain}}

\affiliation[uma]{
            organization={Departamento de Lenguajes y Ciencias de la Computaci\'on, University of M\'alaga},
            city={M\'alaga},
            postcode={29071}, 
            country={Spain}}
            
\affiliation[dasci]{
            organization={Andalusian Research Institute in Data Science and Computational Intelligence (DaSCI)},
            city={C\'ordoba},
            country={Spain}}
\cortext[cor1]{email: jrromero@uco.es}

\begin{abstract}
Automatic workflow composition (AWC) is a relevant problem in automated machine learning (AutoML) that allows finding suitable sequences of preprocessing and prediction models together with their optimal hyperparameters. This problem can be solved using evolutionary algorithms and, in particular, grammar-guided genetic programming (G3P). Current G3P approaches to AWC define a fixed grammar that formally specifies how workflow elements can be combined and which algorithms can be included. In this paper we present \ourmethod, an interactive G3P algorithm that allows users to dynamically modify the grammar to prune the search space and focus on their regions of interest. Our proposal is the first to combine the advantages of a G3P method with ideas from interactive optimisation and human-guided machine learning, an area little explored in the context of AutoML. To evaluate our approach, we present an experimental study in which 20 participants interact with \ourmethod to evolve workflows according to their preferences. Our results confirm that the collaboration between \ourmethod and humans allows us to find high-performance workflows in terms of accuracy that require less tuning time than those found without human intervention.
\end{abstract}




\end{frontmatter}


\section{Introduction}
\label{sec:intro}

Building machine learning (ML) applications require detailed analysis of the raw data, how it should be preprocessed and which algorithms are most effective. Therefore, a strong background in statistics, data visualisation and learning paradigms is necessary to make the right decisions when building the ML workflow. Data scientists acquire these specialised skills after years of training and practice. Other user profiles with less technical ML expertise also interact with the ML workflow and its results. To democratise access to ML, the field of AutoML~\cite{barbudo2023} emerges with the goal of simplifying the definition and use of ML workflows by automating repetitive and time-consuming phases. Primarily conceived to help data scientists~\cite{olson2016}, AutoML also makes the knowledge discovery process more accessible to domain experts~\cite{feurer2015}.

Different tasks can be carried out under the precepts of AutoML, such as neural architecture search~\cite{he2021}, algorithm selection~\cite{luo2016} and hyperparameter optimisation~\cite{yang2020}. However, current AutoML approaches focus mainly on model building~\cite{bilalli2016}, limiting applicability to one step of the knowledge discovery process or even to a particular type of ML model~\cite{liu2023,revin2023}. More recently, authors have started to explore the problem of automatic workflow composition (AWC), with or without hyperparameter optimisation, in the hope of providing more general support for the knowledge discovery process. The AWC problem is usually formulated as an optimisation problem for which solving techniques like Bayesian optimisation~\cite{thornton2013,feurer2015} and evolutionary algorithms~\cite{olson2016, de2017, barbudo2021} are applied. Among evolutionary techniques, those based on genetic programming (GP) are particularly relevant~\cite{olson2016, larcher2019} as workflow structures can be encoded as trees.

Despite advances in AutoML approaches, in general, and AWC, in particular, key challenges to adopting them in practice remain. A recent work has conducted a series of interviews among AutoML users in an attempt to understand the obstacles to implementing AutoML solutions in real-world environments~\cite{sun2023}. Transparency, privacy, and customisation were highlighted as the main challenges. The need to incorporate domain knowledge and tracking the AutoML process were mentioned among possible solutions to increase customisation and improve transparency, respectively. Indeed, both data scientists and domain experts have deep knowledge that could complement or limit the range of possible ML solutions~\cite{gil2019}. However, current AWC methods hide how the returned workflow was found and what alternatives were ruled out in the process. Another important aspect of evolutionary AWC approaches is that they are computationally expensive, as each candidate workflow must be fitted to the dataset. This process require hours or even days if the dataset is large or the workflow includes complex algorithms~\cite{barbudo2021}.

Incorporating user knowledge into the different stages of the ML process is the main objective of human-in-the-loop ML~\cite{mosqueira2023}. This broad term encompasses different degrees of collaboration between humans and ML algorithms, seeking new ways to reach accurate solutions faster and better adapted to user needs. Interactive ML is the approach that fosters closer collaboration between humans and ML algorithms~\cite{amershi2014}. Originally conceived to allow humans to build classifiers interactively~\cite{ware2001}, other tasks such as feature selection~\cite{correia2019}, rule mining~\cite{baum2020}, and clustering~\cite{bae2020} have also benefited from human intervention. Due to the novelty of both fields, the intersection between AutoML and interactive ML has barely been explored. Design principles for human-guided AutoML have been proposed~\cite{gil2019}, specifying where domain experts might incorporate their knowledge to influence how an AutoML system generates ML models. Focusing on AWC, some authors are asking to rethink the role of humans in the process~\cite{lee2020}, as pure automatic workflow might not be adequate in all decision-making scenarios.

To meet users' expectations and requirements for AWC, this paper proposes \ourmethod: a grammar-based interactive genetic programming approach for AWC. We hypothesise that if users are allowed to inspect evolving workflows and provide suggestions, the algorithm will be able to achieve the desired balance between predictive performance and its fitting time. Since \ourmethod is an evolutionary algorithm, we adopt the principles of interactive optimisation methods~\cite{meignan2015} to integrate human feedback across generations. We leverage the flexibility of a context-free grammar (CFG), which defines how workflows can be composed, to interactively adjust the features of candidate workflows. More specifically, \ourmethod allows the human to remove specific algorithms and hyperparameter values from the grammar definition, also deciding how often he/she wants to intervene. As a result of human actions, \ourmethod explores the search space more efficiently, avoiding wasting time optimising workflows that do not satisfy human preferences. To validate our approach, two experiments are presented. A first experiment is conducted under laboratory settings, simulating the interaction of different user profiles. A second experiment with 20 participants provides us with a more realistic scenario to analyse the performance and utility of \ourmethod. To our knowledge, \ourmethod is the first AWC algorithm that supports humans in-the-loop, and one of the few interactive AutoML proposals that performs empirical experiments with humans~\cite{park2020}. Results from simulated users reveal that predictive performance is barely affected, while the reduction in evaluation time is noticeable in several datasets. As for the human experiment, more than 75\% of the participants improved fitness, evaluation time or both compared to the execution of our method without any interaction. In addition, most participants find the method useful and intuitive, as it helps to better understand the generated solutions and to find alternatives not initially conceived.

The rest of the paper is organised as follows. Section~\ref{sec:background} introduces evolutionary workflow composition and interactive machine learning. Related work is summarised in Section~\ref{sec:related}. Section~\ref{sec:method} presents \ourmethod, including the interactive interface built on top of it. The experimental methodology is detailed in Section~\ref{sec:methodology}, while the results are analysed in Section~\ref{sec:results}. Concluding remarks and future lines of research are outlined in Section~\ref{sec:conclusions}.

\section{Background}
\label{sec:background}

\subsection{Evolutionary workflow composition}
\label{subsec:awc}

Automatic workflow composition involves the optimisation of three related problems: 1) which algorithms should be applied; 2) how they are sequenced; and, optionally, 3) what are the values of their hyperparameters. The algorithm selection problem~\cite{rice1976} concerns the recommendation of the best algorithm(s) for a given dataset, and focuses on a specific phase of the knowledge discovery process. The second problem consists in finding the order of execution of the algorithms, since interactions between them could also influence performance. Finally, the hyperparameter optimisation problem~\cite{yang2020} involves selecting the best hyperparameter values for the algorithms, such as the depth of a decision tree or the kernel used by a support vector machine. Initially studied independently, Thornton et al. were the first to formally define the combined problem of algorithm selection and hyperparameter optimisation~\cite{thornton2013}. In general, AWC proposals deal with classification or regression tasks, using a performance metric (\eg accuracy or root mean square error) to guide the optimisation process. Early proposals for classification workflows were based on particle swarm optimisation~\cite{escalante2009} and Bayesian optimisation, such as Auto-WEKA~\cite{thornton2013} and Auto-Sklearn~\cite{feurer2015}. However, most of these proposals predefine the workflow structure, usually composed of one or two preprocessing algorithms followed by a classifier. In the following, we explain how evolutionary algorithms can solve the AWC problem for classification with a more flexible workflow definition.

In evolutionary AWC, candidate workflows evolve through an iterative process. From the set of available algorithms and their possible hyperparameter values, random workflows are initially created. The quality of each workflow is evaluated by training it on the dataset. The algorithm then enters the main loop, where each iteration is called a generation. In each generation, a four-step process is applied: parent selection, to choose some workflows for reproduction; crossover, where the selected parents are recombined to generate new workflows; mutation, to introduce some diversity by altering the structure or hyperparameter values of some workflow algorithms; and replacement, which consists of selecting those workflows that will survive for the next generation. A special type of evolutionary algorithm is genetic programming (GP), a paradigm that represents solutions as trees to evolve programs. GP has often been used to induce classifiers~\cite{rivera2022}, including feature selection if necessary~\cite{muni2006}. In addition, its features make GP a highly configurable approach to control the complexity of the resulting models, being inherently interpretable~\cite{mei2023}. All of these ideas can be extrapolated to AWC, where trees represent fully functional workflows rather than a single model.

TPOT~\citep{olson2016} was the first GP approach to building classification workflows under a multi-objective perspective. TPOT supports multi-branch workflows and is designed to maximise the workflow predictive performance while minimising its size. To mitigate the shortcoming of costly workflow evaluations, TPOT has been extended to first fit the workflows with a subset of the data~\citep{gijsbers2017, parmentier2019}. This way, the full dataset is only used to evaluate the most promising workflows. To avoid generating invalid workflows, some recent proposals have explored the use of grammar-oriented approaches, such as grammar-guided genetic programming (G3P) and grammatical evolution. RECIPE is a G3P approach whose grammar defines how different types of preprocessing algorithms should be ordered~\cite{de2017}. Also based on G3P, Auto-CVE~\cite{larcher2019} uses the grammar to specify that workflows should follow a specific sequence of three algorithms: preprocessing, feature selection, and classifier. Preprocessing and feature selection are optional. Based on the same three-step workflow structure, HML-Opt~\cite{estevez2020} applies grammatical evolution. In this case, more than one algorithm from each category can appear in each step thanks to a recursive composition process. AutoML-DSGE~\cite{assunccao2020} is another grammar evolution approach to build scikit-learn pipelines. This method adopts the RECIPE grammar, but customises it for each dataset.

\subsection{Interactive machine learning}
\label{subsec:interactive-ml}

Interactive ML is part of the so-called human-centred ML~\cite{mosqueira2023}, a broad area that encourages collaboration between humans and ML. The main idea behind interactive ML is to allow humans to provide feedback to an ML technique, so that the learning process takes human preferences into account. As a result of integrating feedback, ML techniques adapt incrementally and provide more focused and useful solutions compared to traditional learning. Humans can be involved in different ML tasks, at different points in the knowledge discovery process, and take on different roles~\cite{ramos2020}. For example, they can be involved in data preparation, such as image annotation, to identify relevant concepts that a machine cannot easily handle~\cite{porter2013}. However, the central activity in interactive ML is monitoring the model building process to add constraints and validate the results~\cite{fails2003}.

According to Dudley and Kristensson~\cite{dudley2018}, an interactive ML system can be described in terms of four elements: users, model, data, and interface. Data and model are intrinsically linked to any ML approach, while users and interface become particularly relevant in interactive ML. The role of the user refers not only to how they interact with the ML system, but also to their level of expertise in relation to ML (data scientists, domain experts, etc.). On the other hand, the design of a user interface is another crucial feature of interactive ML proposals. The interface enables interaction options and can greatly influence how the human understands the task to be performed~\cite{mosqueira2023}.

How human feedback is collected and integrated depends largely on the ML task in which the human-machine collaboration is framed. Since our work uses an optimisation algorithm (G3P) to address the AutoML task (AWC), we will focus on how interactive optimisation works. In this sense, Meignan et al.~\cite{meignan2015} identify five main types of interactive approaches: ``trial and error'', which actually represents running the algorithm multiple times with some adjustment (\eg changing parameter values); interactive reoptimisation, in which the user redefines the problem formulation by introducing changes to an intermediate solution; interactive multiobjective optimisation, suitable when the user seeks to dynamically explore trade-offs between objectives; interactive evolutionary algorithm, which follows the traditional idea of replacing the fitness function with a human subjective evaluation; and human-guided search, which allows the user to modify parts of the solutions or the algorithm's behaviour to guide the search. To fully characterise an interactive optimisation algorithm, other aspects need to be designed~\cite{meignan2015,shackelford2007}:

\begin{itemize}
    \item \emph{Selection of solutions}. It involves deciding how many solutions from the population will be shown to the user and how they are selected.
    \item \emph{Type of user feedback}. It specifies the mechanism offered by the algorithm to provide feedback. For example, the user may be asked to score solutions, set weights for objectives, or mark parts of solutions that need to be modified.
    \item \emph{Interaction schema}. It includes the design of when the interaction starts and how often it occurs. These decisions can be fixed (every N generations) or adapted based on the search progress.
    \item \emph{Preference information lifetime}. It refers to the time during which the information is considered valid, and whether or not the user's decisions can be reversed.
\end{itemize}

\section{Related work}
\label{sec:related}

This section discusses current attempts to add some degree of interactivity in AutoML, with special attention to the AWC problem. However, it should be noted that most of the proposals (and all AWC-oriented ones) do not support human intervention beyond setting some configuration options and visualising the results. Therefore, these works should be classified as ``trial-and-error'' approaches, as defined in Section~\ref{subsec:interactive-ml}. Furthermore, none of these proposals rely on evolutionary algorithms.

Gil et al. have proposed a conceptual framework for defining AutoML tasks that could benefit from human collaboration~\cite{gil2019}. The tasks summarise preferences the user may have regarding data (feature and instance selection), model development (specific models or hyperparameters are desired) and evaluation (preferred metrics for model comparison). For each task, the authors describe the reasons for user interaction and the components of the AutoML system that would be affected. Lee et al. have conducted a similar study but oriented to the AWC problem~\cite{lee2020}. They identify three possible levels of autonomy: user-driven, cruise-control and autopilot. At the user-driven level, the user can make all modelling decisions to specify how the workflow should be created. The cruise-control level encourages further automation, allowing less expert users to only set their workflow requirements and react to the solutions proposed by the system. Finally, the autopilot level corresponds to the traditional AWC approach, where the user only specifies the dataset and the type of problem, \eg classification or regression.

Some AutoML visual tools include options for configuring the workflow composition process and inspecting the resulting workflows. QuIC-M is a data exploration and workflow composition tool that allows the user interleave both tasks by adding constraints to model building~\cite{binning2018}. The user can specify which parts of the workflow should not be modified or change the evaluation to prioritise runtime. The tool designed by Elkholy et al.~\cite{elkholy2019} sequentially assists the user with the three AWC-related problems: algorithm selection, workflow composition, and hyperparameter optimisation. The user determines the configuration options before each step is executed, thus he/she can inspect the intermediate results. PipelineProfiler supports the analysis of workflows generated by different AutoML techniques~\cite{ono2020}. Among its main features, the user can analyse the relationship between workflow elements and an evaluation metric (\eg execution time or accuracy).

AutoDS is a prototype system that helps users perform ML tasks~\cite{wang2021}. Once the user loads the dataset, AutoDS proposes the configuration of a classification or regression task. The user must approve the configuration or adjust it. AutoDS then runs autonomously, though the user can monitor the execution in real time. The resulting workflows are presented visually as trees, along with performance metrics. The authors conducted a user study with two groups of participants, one using AutoDS and one unsupported by the tool. Both groups were asked to build classification workflows in order to compare solutions manually crafted and those generated by the tool.

\section{Description of \ourmethod}
\label{sec:method}

\subsection{The evolutionary search process}\label{subsec:method-overview}

\begin{figure}[!t]
\centerline{\includegraphics[width=0.62\textwidth]{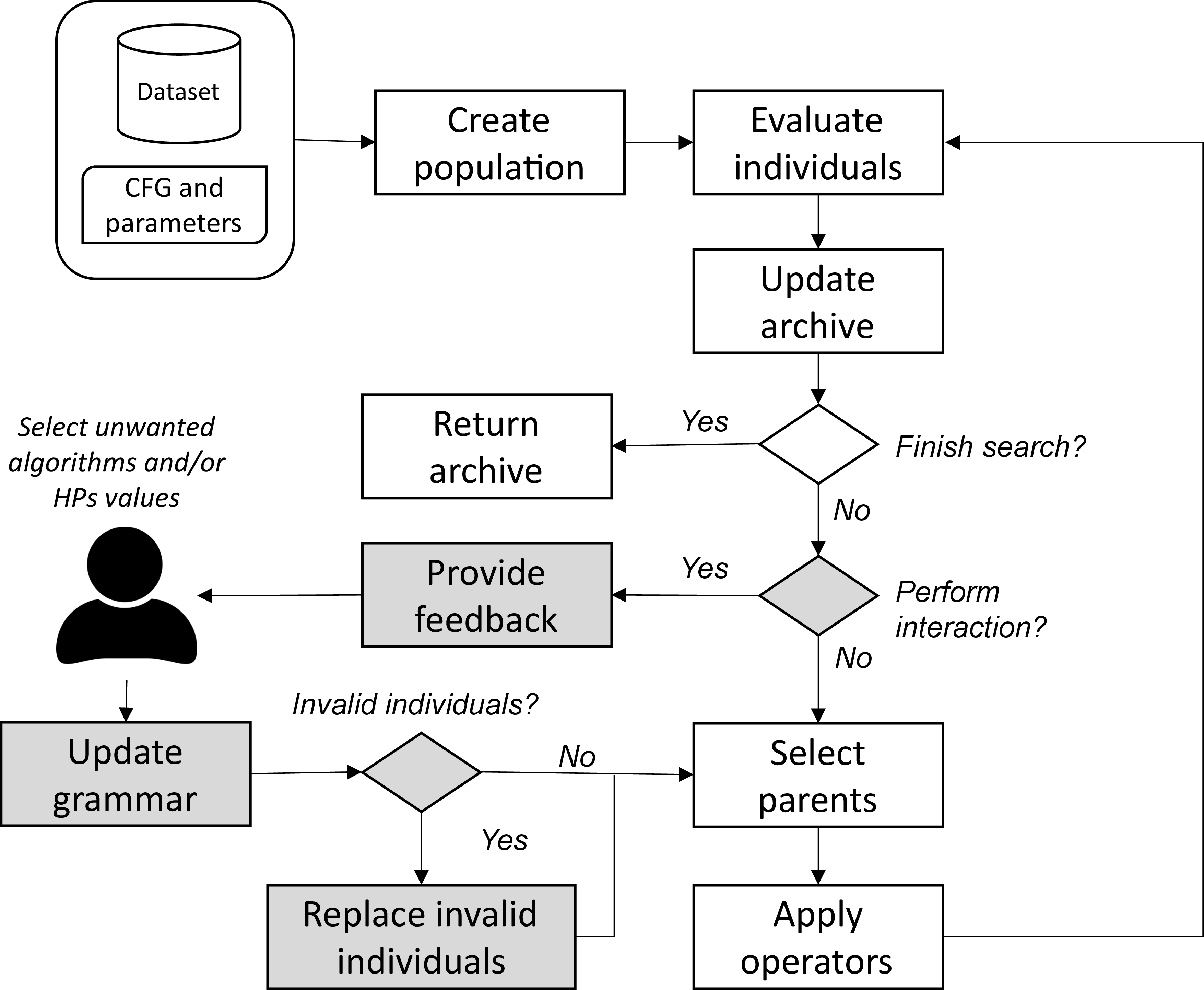}}
\caption{Proposed interactive evolutionary model.}
\label{fig:model}
\end{figure}

Figure~\ref{fig:model} shows the general procedure of \ourmethod, which receives two main inputs: a context-free grammar that defines the structure of any valid individual (\ie workflow) and a classification dataset that is used to evaluate the candidate workflows. Other inputs are: the maximum number of generations, the maximum number of interactions, the population size, the crossover and mutation probabilities, and the number of generations to run before the first interaction. The evolutionary process starts by randomly generating a population of workflows according to the CFG. Then, the workflows are evaluated using the fitness function, which trains the workflow for the input dataset and computes the balanced accuracy. Specifically, a 5-fold cross validation is adopted to mitigate the risk of overfitting. The best individual is then stored in an archive to ensure that it is not lost as a result of human interaction.

Once the archive has been updated, \ourmethod checks whether the optimisation process should continue or not. If the maximum number of generations or interactions has been reached, the execution ends and the archive is returned. Otherwise, \ourmethod checks whether the current generation requires human interaction or not. As detailed in Section~\ref{subsec:interactive-mechanisms}, human interactions are scheduled at the user's discretion, with the exception of the first one which is set to occur after the first $N$ generations have been completed (we set $N=15$ in our experiments). If there is no scheduled interaction in the current generation, the genetic operators are applied. First, the selection operator chooses the parents to be used for breeding. To do this, \ourmethod implements a binary tournament. Next, offspring are generated by applying the crossover and mutation operators based on their probabilities. On the one hand, the crossover operator randomly selects two branches (one per parent) with the same non-terminal root and swaps them. On the other hand, the mutation operator chooses a non-terminal at random, which is deleted, and reconstruct the corresponding tree branch. Finally, offspring are evaluated, the archive is updated and the offspring replace the former population ensuring that the best workflow is kept.

If the current generation involves human interaction, \ourmethod will provide the human with information on the state of the population and the optimisation process. Based on the information provided, the human can modify the search space by removing algorithms and/or hyperparameter values from the CFG. Notice that this may cause some workflows from the current population to be invalid, as they do not satisfy the constraints imposed by the modified grammar. Therefore, these workflows are removed from the population and new randomly generated individuals are introduced to keep the population size constant. The new individuals, which comply with the modified grammar, are then evaluated. The evolutionary process can be resumed at this point, whereby the genetic operators are applied and evolution continues to the next generation.

\subsection{Interactive approach}
\label{subsec:interactive-mechanisms}

Our interactive G3P algorithm follows the idea of the reoptimisation approach described in Section~\ref{subsec:interactive-ml}, as the main purpose of the human intervention is to redefine the problem by adjusting the CFG. The user action is focused on the removal of  algorithms and/or hyperparameter values during the interactions, so that the candidate workflows only present those algorithms and hyperparameter values the human is interested in. Focusing on the algorithms, the human has the freedom to discard any of them, no matter if the algorithm performs a preprocessing or a classification operation. The only restriction is that the CFG should kept at least one classification algorithm, otherwise no valid workflows can be derived. As for the hyperparameter values, the user can delete concrete values of categorical hyperparameters. The evolutionary algorithm is the only responsible for the optimisation of the numerical hyperparameters, since it would be rather irrelevant to discard a single value. As happens with classification algorithms, the CFG should include at least one categorical value for each hyperparameter.

Whenever the human decides to discard an algorithm or hyperparameter value, \ourmethod has to delete it from the CFG.\footnote{The full grammar, together with a list describing the available algorithms and hyperparameter values, is detailed in the additional material.} Preprocessing algorithms are obtained from the imbalanced-learn and scikit-learn libraries, whereas classification algorithms are taken from scikit-learn and XGBoost libraries. The \ourmethod grammar consists of a set of terminal symbols ($\sum_T$), a set of non-terminal symbols ($\sum_N$), and the production rules (\textit{P}) for deriving expressions from the root symbol (\textit{P}). Terminal symbols represent the algorithms, together with their respective hyperparameters and supported values. Non-terminal symbols define the derivable elements to represent a sequence of algorithms in the workflow and how each one is configured. Production rules determine the derivation steps required to generate valid workflows, which are ultimately written in terms of terminal symbols, \ie algorithms and hyperparameters. When a symbol of the grammar has to be removed as a result of human decision, the grammar undergoes modifications at different levels. If an algorithm (\eg decision tree) has been chosen to be removed, the symbols representing the algorithm  (\textit{decisionTree}) and its hyperparameters (\textit{$<$decisionTree$\_$hp$>$}, \textit{criterion}, \textit{maxDepth}, etc.) are deleted. This implies that the algorithm and all its hyperparameter values no longer belong to the CFG. Furthermore, \ourmethod must forget the production rule that derives the \textit{<classifier>} symbol in that specific classifier and the rule that derives \textit{$<$decisionTree$\_$hp$>$} in the set of hyperparameters (\ie \textit{criterion}, \textit{maxDepth}, etc.). If the symbol to be removed is a hyperparameter value (\eg \textit{gini} value for \textit{criterion}), \ourmethod has to modify the function associated to the hyperparameter \textit{criterion}. For the rest of the evolution, if the hyperparameter symbol appears on an individual, the function will randomly return one of the values that are still valid for such hyperparameter. The rest of aspects that characterise our interactive approach are defined below:

\paragraph{Selection of solutions} In each interaction, the human inspects all individuals that have been evaluated since the last interaction. In case no interaction has occurred yet, \ourmethod selects all the individuals evaluated so far. Focusing on the information shown, the workflow represented by the individual is visualised in textual format, accompanied by its fitness value (balanced accuracy) and evaluation time. We also indicate the best workflow in the current population and the best global individual of the whole optimisation process. The best workflow is the one with the highest fitness value and, in case of a tie, the minimum evaluation time. 

\paragraph{Type of user feedback} Although humans can see the evaluation time, fitness and algorithm sequence for each workflow, information on the state of the population is provided in a more synthesised form. For each algorithm and hyperparameter value, the maximum and average evaluation time, and the maximum fitness achieved by the workflows containing them are displayed. To better contextualise these statistics, we also calculate how many times each algorithm/hyperparameter value appears in the set of individuals shown. Consider the case where the human decides to eliminate a classifier that is not accurate. However, his/her intuition could be wrong if the classifier only appears in one workflow, as these poor results could be due to the preprocessing sequence.

Based on the intermediate results, feedback is provided in two ways. First, the human must set the values of two thresholds: $t_{acc}$, to indicate the minimum accuracy desired, and $t_{time}$, to limit the maximum evaluation time. The intersection of these two thresholds divides the workflow search space into two regions: $R_{best}$, with the workflows that have greater accuracy than $t_{acc}$ and lower evaluation time than $t_{time}$; and $R_{worst}$, with the workflows in which the human is not interested because they have either lower accuracy than $t_{acc}$ or higher evaluation time than $t_{time}$. Only those algorithms and hyperparameter values that appear in workflows belonging to $R_{worst}$ and do not appear in any workflow belonging to $R_{best}$ are candidate elements for removal from the CFG. Once the candidate algorithms and hyperparameter values are isolated, the human specifies whether or not to discard some of them. Based on the human choice, \ourmethod will receive the list of grammar symbols to be eliminated and perform the grammar modification.

\paragraph{Interaction schema} The scheduling of interactions throughout the search process is controlled by the maximum number of generations and the maximum number of interactions. Once execution has started, \ourmethod runs until it reaches the generation at which the first interaction was configured. From this point on, the human can specify how many generations \ourmethod must execute until the next interaction. The human is informed about how many interactions and generations are still available. At any given interaction, the human also decides whether to continue with the interactive optimisation process or not. Therefore, we provide a great deal of flexibility to the human as to how many times and when he/she will interact.

\paragraph{Preference information lifetime} Human decisions are valid for the remaining generations, \ie the removal of algorithms and hyperparameter values is applied until the end of the search. This means that individuals in the current population that include the removed elements are discarded, and will only remain if they appear among the best individuals stored in the archive. The grammar modification is permanent, \ie it cannot be reversed, so the evolutionary algorithm will not generate any new workflow with the removed algorithms or hyperparameter values. Letting the user undo the grammar modification may cause the algorithm to lose convergence as the search space expands again. Therefore, we prefer to guide \ourmethod to smaller regions of the search space after each interaction.

\subsection{Interactive interface}
\label{subsec:tool}

\begin{figure}[!t]
\centerline{\includegraphics[width=\textwidth]{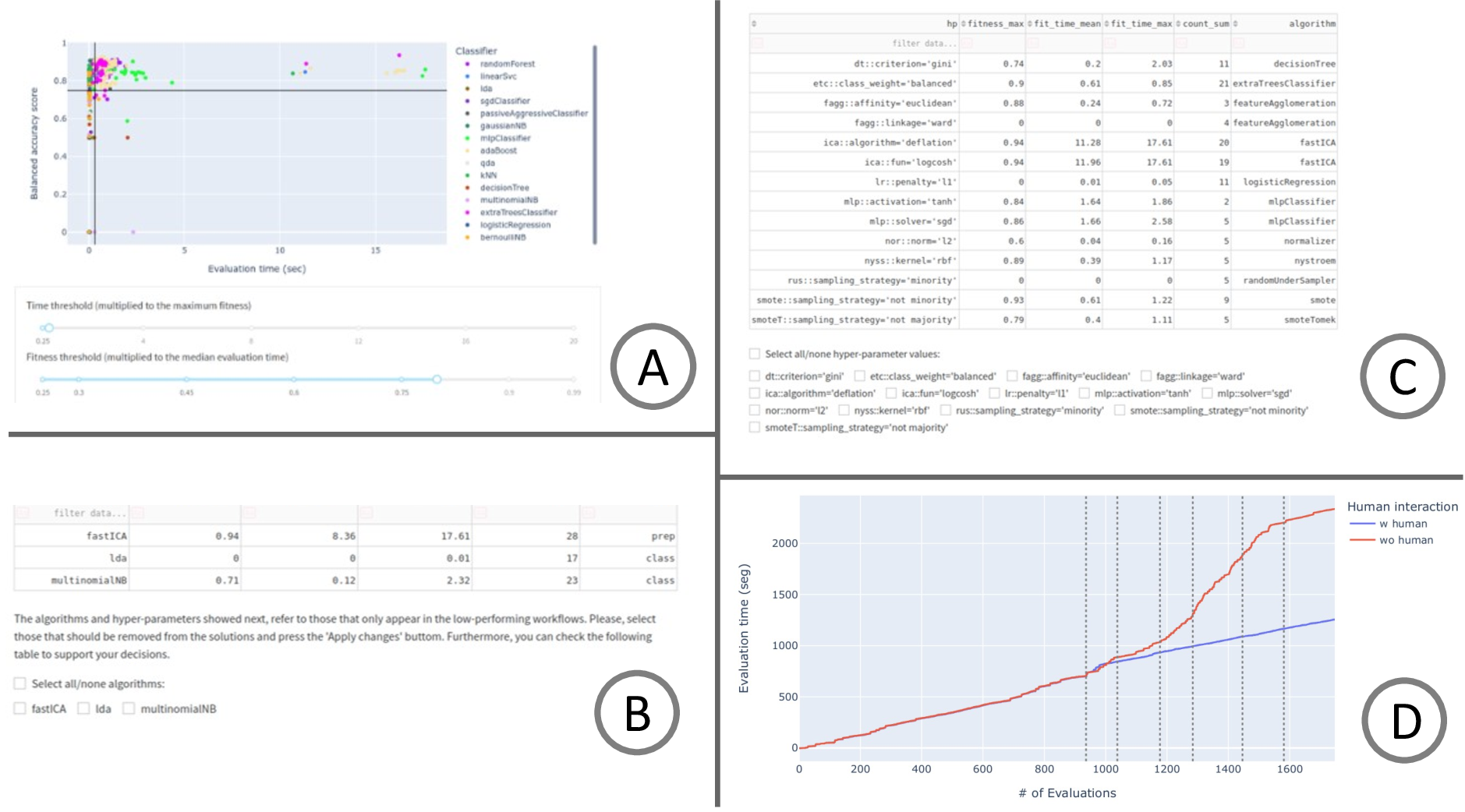}}
\caption{Graphical user interface of \ourmethod.}
\label{fig:tool-gui}
\end{figure}

We have developed a web-based graphical user interface to support the interactive process and conduct our experimentation. The interface provides the information of the evolutionary process in multiple visual formats. Figure~\ref{fig:tool-gui}A shows the scatter plot used to visualise all individuals evaluated since the last interaction. The x-axis symbolises the evaluation time in seconds, while the y-axis represents the fitness, \ie the balanced accuracy. In addition, the colour of the dot indicates the classifier that appears at the end of the workflow. The complete sequence of algorithms is visible by hovering the mouse over a particular point. The graphic is interactive, allowing the user to zoom in and hide or show the workflows containing certain classifiers by clicking on the legend. The black horizontal and vertical lines mark the thresholds used to separate the $R_{best}$ and $R_{worst}$ regions. In this way, the human can clearly isolate the workflows that fall into each region. The thresholds can be adjusted with the sliders at the bottom of the scatter plot.

Based on $R_{best}$ and $R_{worst}$, the interface populates two tables with the statistics of the algorithms (Figure~\ref{fig:tool-gui}B) and hyperparameter values (Figure~\ref{fig:tool-gui}C) that can be deleted. Apart from the statistics explained in Section~\ref{subsec:interactive-mechanisms}, the table with the algorithms indicates the algorithm type (preprocessing or classification). In the table summarising the hyperparameter values, the last column indicates to which algorithm the hyperparameter belongs. Below these tables, the interface activates two checklists with the algorithms and hyperparameter values so that the human can select which ones will be actually removed.

The removal of certain algorithms and hyperparameter values can greatly influence the workflow evaluation time and, consequently, the execution time of the evolutionary algorithm. To illustrate it, the interface includes a line plot (Figure~\ref{fig:tool-gui}D) to show how the execution time diverges from the baseline ---50 generations without human intervention--- as the human interacts.

\section{Experimental methodology}
\label{sec:methodology}

\subsection{Research questions}\label{subsec:methodoloy-rq}

Three research questions guide the experimental evaluation of \ourmethod:

\begin{itemize}
    \item \textbf{RQ1:} \emph{How does the interactive adaptation of the CFG influence the performance of \ourmethod?} The modification of the terminals and production rules of the CFG implies the redefinition of the search space after each interaction. We need to investigate how this change affect the behaviour of the algorithm and its impact on the quality of the workflows in terms of predictive accuracy and evaluation time.
    
    \item \textbf{RQ2:} \emph{Which algorithms and hyperparameter values are more frequently discarded by humans during the interaction?} Users may have different preferences regarding the exploration of candidate workflows. Therefore, we want to analyse which elements of the CFG they prefer to eliminate. In addition, each participant may be more or less conservative in the number of modifications to the GFC as interactions progress.
    
    \item \textbf{RQ3:} \emph{How does human intervention impact the trade-off between accuracy and evaluation time?} The decisions made by the participants possibly respond to different causes, as some of them might prioritise accuracy over evaluation time or vice versa. Also, we seek to understand how the executions of \ourmethod guided by the participants differ from a baseline execution in which no interaction happens.
    
    \item \textbf{RQ4:} \emph{Which aspects related to the user experience do the participants highlight the most?} As part of the user study, we gather feedback regarding the use of an interactive tool to address the AWC problem. The qualitative analysis of the participants' opinions after the experiment allows us to discuss the benefits of \ourmethod.
\end{itemize}

To respond to these RQs, two experiments are conducted. First, we need to analyse how the algorithm reacts to changes in the grammar definition for a variety of datasets and user actions. Therefore, we run the algorithm in a laboratory environment simulating human interactions. The methodology for carrying out this experiment, which is aligned to RQ1, is explained in Section~\ref{subsec:methodology-experiment1}. \ourmethod is then evaluated in a realistic scenario based on a user study with 20 participants. As detailed in Section~\ref{subsec:methodology-experiment2}, participants were asked to perform a run with the interactive tool for a specific dataset. The logs generated by \ourmethod and the questionnaire completed by the participants provide us with the necessary information to answer RQ2-RQ4. Combining simulated and real interactions is a common approach in interactive optimisation to cover a wider range of experimental conditions~\citep{marculescu2015,ramirez2018}.

\subsection{Datasets and parameter set-up}\label{subsec:methodoloy-datasets}

\begin{table}[!t]
\centering
\scalebox{0.9}{
\begin{tabular}{lcccc}
\hline
 & Features & Classes & Train & Test \\
\hline
breastcancer* & 9 & 2 & 466 & 233 \\
germancredit &20 &2 &700 &300 \\
glass* & 9 & 6 & 142 & 72 \\
hillvalley* & 100 & 2 & 808 & 404 \\
ionosphere* & 33 & 2 & 234 & 117 \\
semeion &256 &10 &1116 &477 \\
winequalityred* & 11 & 6 & 1066 & 533 \\
winequalitywhite &11 &7 &3429 &1469 \\
yeast &8 &10 &1039 &445 \\
\hline
\end{tabular}
}
\caption{Datasets used for experiments}
\label{tab:datasets}
\end{table}

For experimentation, we consider nine classification datasets used in other AWC studies. Table~\ref{tab:datasets} summarises the characteristics of each dataset: the number of features and classes, and the size of the training and test sets. For the datasets taken from~\citep{thornton2013}, we apply the same data partition defined by the authors. The datasets taken from~\citep{de2017}, which are marked with an asterisk (*), are randomly partitioned with one-third of the samples used for validation to maintain consistent conditions across all datasets.

The parameter values used by the G3P algorithm that are common to both experiments are: population size (100), crossover probability (0.8), mutation probability (0.2) and maximum number of derivations (13). The population size was chosen to avoid human experimentation taking excessive time. As for the crossover and mutation probabilities, they were determined by preliminary experiments. The maximum number of derivations was set to 13, which corresponds to workflows composed of up to five elements.

\subsection{Experiment 1: Laboratory settings}
\label{subsec:methodology-experiment1}

The goal of this experiment is to evaluate how well \ourmethod adapts to changes in the CFG definition. More specifically, we aim to determine how the algorithm behaves in comparison to an evolutionary process without interaction. To draw generalisable conclusions, we need to run \ourmethod for a variety of datasets and grammar modifications, also taking into account the random nature of evolutionary algorithms. In this scenario, performing the experiment with human participants would be overwhelming in terms of cognitive load and time required. Therefore, we simulate human interactions by defining different user profiles that will apply changes to the grammar at fixed times during the search. Each profile will be characterised by a set of predefined rules that specify the user's preferences as to which algorithms and hyperparameter values of the grammar should be removed. In this way, we can simulate users who place more importance on achieving high predictive accuracy, while others will prioritise workflows with low evaluation time.

Each user profile will follow a different strategy to modify the CFG depending on three aspects: 1) the importance given to the workflow predictive accuracy, 2) the importance given to the workflow evaluation time, and 3) how disruptive the user is in choosing the elements to remove. Based on criteria (1) and (2), we will assign different values to the $t_{acc}$ and $t_{time}$ thresholds for each user profile (see Section~\ref{subsec:interactive-mechanisms}). 

\begin{table}[!t]
\centering
\scalebox{0.9}{
\begin{tabular}{ll} 
\hline
Criterion & Alternatives \\ 
\hline
Accuracy threshold ($t_{acc}$)          & [0, 0.8, 0.9] x average fitness \\
Evaluation time threshold ($t_{time}$)  & [0, 0.5, 1] x median evaluation time\\
Algorithm removal strategy  & Most frequent algorithm\\
                            & Max. 1/3 of total number of algorithms\\
Hyperparameter removal strategy & All possible hyperparameter values\\
\hline
\end{tabular}
}
\caption{Criteria to define the profiles of simulated users.}
\label{tab:threholds}
\end{table}

To establish $t_{acc}$, we calculate the average fitness of the individuals evaluated since the previous interaction. For the first interaction, all individuals evaluated up to that point are considered. Then, the threshold is obtained as the average fitness multiplied by a fixed constant that will vary depending on the user profile. Similarly, $t_{time}$ is set as a function of the median evaluation time of the individuals generated since the last interaction, also multiplied by a fixed constant. We decided to use these relative thresholds because the accuracy and evaluation time can vary greatly depending on the dataset. Thus, the thresholds can be automatically adjusted depending on the progress of the evolution. Table~\ref{tab:threholds} shows the three constant values used for each threshold, resulting in nine combinations. However, the combination $t_{acc}=0$ and $t_{time}=0$ is discarded as it would create a $R_{worst}$ region with no algorithm or hyperparameter to remove.

Once the region $R_{worst}$ is isolated, we have to decide how the simulated user would act, \ie how many algorithms and hyperparameter values will be removed from the CFG. We have defined two alternatives, as summarised in Table~\ref{tab:threholds}. Focusing on algorithms, the user can remove only one algorithm, the one that appears the most times in $R_{worst}$. Alternatively, we include a less conservative option: a maximum of 1/3 of the algorithms originally defined in the CFG can be eliminated by choosing again those that appear most frequently. We set such a maximum to avoid eliminating all algorithms in successive interactions. As for the hyperparameter values, since they represent smaller changes in the workflow, we allow removing all hyperparameter values in $R_{worst}$. In total, we define 16 different user profiles after combining thresholds and removal policy options.

Once the user profiles have been created, the interaction schedule must be defined. Since user decisions will be made automatically, the time at which the user intervenes and the total number of interactions must also be specified prior to execution. Considering that 1/3 of the algorithms and all hyperparameter values can be eliminated in one interaction, the total number of interactions cannot be too high. We decided to set two interaction moments during the 50 generations as follows. The algorithm is allowed to run 15 generations without any interaction. Thresholds are then determined based on the state of the population and the first user interaction is simulated. The algorithm is run for another 15 generations, performing a new interaction with updated thresholds as the population has changed. After the second interaction, the population evolves for the last 20 generations. This type of interactive scheme allows the algorithm to provide sufficiently good solutions in the initial interaction, while letting the algorithm incorporate the last preferences before concluding the search~\cite{ramirez2018}. 

In this experiment, \ourmethod is run for all datasets included in Table~\ref{tab:datasets}, setting 50 generations as the stopping condition. 30 runs are executed for each simulated user profile (16 in total) and dataset (9). In total, this represents 4050 runs of \ourmethod, which would not be achievable with human participants. In addition, we run \ourmethod without any interaction enabled for each dataset (30 repetitions) to have a baseline to compare against.

\subsection{Experiment 2: Human interaction}
\label{subsec:methodology-experiment2}

In this experiment, participants interact with \ourmethod to discard algorithms and hyperparameter values. Next we describe the organisation of the experiment, the selection of participants and the data collection process.

\paragraph{Organisation of the experiment}

The interactive experiment is organised in three parts: 1) a brief introduction to the context of the experiment; 2) a demonstration run using the interaction interface to familiarise participants with it; and 3) the actual interactive run with \ourmethod. Although the participant can interrupt the third part at any time, we plan for the session to last less than 1 hour and 30 minutes to avoid fatigue.

For the demonstration, we chose a small dataset (iris) which is also quite popular among ML students and practitioners. For the actual interactive experiment, all participants apply the same dataset (ionosphere) and use the same random seed. In this way, all participants start from the same initial population and we can compare how their different behaviours influence the evolution of the search process. The selection of the dataset and the random seed was carefully made on the basis of two criteria. A first criterion was the total time required by \ourmethod to run the 50 generations. To avoid fatigue and maintain the participant's interest, we excluded those datasets whose AWC process was computationally expensive according to Experiment 1. Our intention was that the interactive execution should not exceed 40-50 minutes. As a secondary criterion, we discarded those particular datasets and random seeds for which \ourmethod showed high convergence in the first generations. Otherwise, the actual human contribution would be substantially reduced.

The population state at the $15^{th}$ generation was serialised, so that the participant does not have to wait to conduct the first interaction. After each interaction, the participant must specify whether he/she wants to continue for some generations or stop the search because he/she is satisfied with the results. If the participant decides to continue, he specifies how many generations are to be run until the next interaction. In total, the participant can schedule 10 interactions over the remaining 35 generations. It should be noted that a version of \ourmethod without any interaction (50 generations) was previously run for reference purposes. It not only allows us to compare the final results, but also to illustrate how the cumulative evaluation time diverges from the participant's execution.

\paragraph{Participants}

We recruited 20 participants with different profiles: three undergraduate students (final year), six PhD students, four assistant/associate professors, one postdoc and four data scientists working in industry (one of them is also a PhD student).\footnote{Due to space limitations, the table summarising the profile and expertise of the participants is provided as part of the additional material.} 

As part of the questionnaire, participants have to rate their knowledge and experience on a 5 Likert scale. The vast majority of participants are very familiar with the algorithms used for preprocessing and classification, and use scikit-learn or similar ML libraries. Such responses give us confidence about their level of understanding of the scikit-learn algorithms and hyperparameter values included in the CFG. As for the research fields involved in this experiment, the participants acknowledge a high knowledge of evolutionary computation, useful for handling the distribution of generations. Participants indicate less knowledge about AWC and AutoML in general. However, it is interesting as \ourmethod aims to bring AWC (and AutoML) closer to humans by helping them in the process. In terms of practical experience in ML, we have different profiles ---from less than one year to more than 10--- to analyse how \ourmethod reacts to various strategies. Although most of the participants have knowledge and practical experience working with evolutionary algorithms, only 25\% of them have had previous experience with interactive approaches. In all cases, they were participants of interactive experiments we conducted in the past for other application domains, being the first time using an interactive method for AWC.

\paragraph{Data collection and questionnaire}
To address RQ2-RQ4, we need to collect information about the participants' actions and decisions. After each interaction, our tool stores the following information:

\begin{itemize}
    \item The time spent by the participant in the interaction.
    \item The list of algorithms and hyperparameter values the participant has chosen to remove from the CFG.
    \item The values of the $t_{acc}$ and $t_{time}$ thresholds that the participant has applied to split the $R_{best}$ and $R_{worst}$ regions.
\end{itemize}

Apart from the automatically generated logs, we provide participants with a questionnaire divided into the following parts:

\begin{enumerate}
    \item \emph{Pre-interaction survey}. This part has served us to extract statistics about the participants' background and experience.
    \item \emph{Interaction survey}. After each interaction, the participant has to justify the reasons for his/her actions. In the last interaction, the participant has to explain why he/she has decided to stop interacting.
    \item \emph{Post-interaction survey}. Once participants have completed the interactive run, we ask them to fill in a survey to assess their level of agreement with some questions. (see Table~\ref{tab:questions}). The first group (Q1.1-Q.18) collects opinions on the whole interaction process. The second group (Q2.1-Q2.4) focuses on the observed evolution of the process throughout the interactions. The third group (Q3.1-Q3.7) is related to the user experience and the practicality of the approach. All the above questions can be rated on a 5 Likert scale ranging from 1=Strongly Disagree to 5=Strongly Agree. Questions Q4.1-4.8 aim to assess the usefulness of the interactive interface elements, for which the participant can give a score from 1=Not useful at all to 5=Very useful. Similarly, questions Q5.1-5.6 allow the user to evaluate the intuitiveness of each option available in the interactive interface. The score should be a value between 1=Not very intuitive and 5=Very intuitive.
\end{enumerate}

\begin{table}[!t]
\centering
\scalebox{0.7}{
\begin{tabular}{l|p{18cm}} 
\hline
\textbf{ID} & \textbf{Question}\\
\hline
Q1.1 & I have found the interactions tedious and time-consuming.\\
Q1.2 & I have paid more attention to the early interactions than the later ones.\\
Q1.3 & The interactive scatter plot has helped me to make a decision.\\
Q1.4 & The tabular information shown has helped me to make a decision.\\
Q1.5 & The time comparison line plot has helped me to make a decision.\\
Q1.6 & I found the interaction options to be appropriate.\\
Q1.7 & I have used the delete algorithm option more frequently than the delete hyperparameter value option.\\
Q1.8 & I have opted more often for eliminating classification algorithms than preprocessing algorithms.\\
\hline
Q2.1 & The algorithm has worked as expected.\\
Q2.2 & I was pleasantly surprised by some of the solutions.\\
Q2.3 & I perceived that my interactions influenced the composition of the workflows.\\
Q2.4 & I perceived how my interactions influenced the evaluation time of the workflows.\\
\hline
Q3.1 & I find it interesting to have an interactive assistant tool for the composition of ML workflows.\\
Q3.2 & I found the information displayed at all times useful.\\
Q3.3 & I have missed more information on how to remove certain algorithms or hyperparameter values.\\
Q3.4 & The time spent on interactions has compensated for the outcome.\\
Q3.5 & The experience has helped me to learn about possible solutions.\\
Q3.6 & I would try \ourmethod with another dataset.\\
Q3.7 & I would apply \ourmethod to my own ML problem.\\
\hline
Q4.1 & Rate the utility of the interactive scatter plot (accuracy vs. evaluation time). \\
Q4.2 & Rate the utility of the column ``Mean evaluation time'' per algorithm in the summary table. \\
Q4.3 & Rate the utility of the column ``Max fitness'' per algorithm in the summary table.\\
Q4.4 & Rate the utility of the ``Mean evaluation time'' per hyperparameter value in the summary table.\\
Q4.5 & Rate the utility of the ``Max fitness'' per hyperparameter value in the summary table.\\
Q4.6 & Rate the utility of the line plot showing the evaluation time with and without human interaction.\\
Q4.7 & Rate the utility of the option to specify how many generations will be executed after the interaction.\\
Q4.8 & Rate the utility of the option to stop the search.\\
\hline
Q5.1 & Rate the intuitiveness of the option to distribute the number of generations and interactions.\\
Q5.2 & Rate the intuitiveness of the options to manipulate the interactive scatter plot.\\
Q5.3 & Rate the intuitiveness of modifying the fitness and time thresholds.\\
Q5.4 & Rate the intuitiveness of the option to delete algorithms.\\
Q5.5 & Rate the intuitiveness of the option to delete hyperparameter values.\\
Q5.6 & Rate the intuitiveness of the option to stop the search.\\
\hline
\end{tabular}
}
\caption{Post-interaction questions in the questionnaire for participants.}
\label{tab:questions}
\end{table}

\section{Experimental results}
\label{sec:results}

\subsection{Experiment 1: Laboratory settings}
\label{subsec:results-exp1}

To answer RQ1, we analyse the fitness and evaluation time of the solutions obtained with the simulated interactions. We adopt the notation $<$$fx\_ty\_az$$>$ to identify each user profile, where $f$, $t$ and $a$ denote fitness, time and algorithm removal strategy, respectively. Note that $f$ or $t$ are omitted when only one threshold is applied. The values $x$, $y$ and $z$ correspond to the thresholds or strategy defined in Table~\ref{tab:threholds}. In terms of fitness, the differences between the execution with and without interaction are minimal. The baseline obtains an average fitness of 0.7750 for the nine datasets, while the solutions found by the simulated users vary between 0.7712 and 0.7755.\footnote{Results by user profile and dataset are available as part of the additional material.} A Friedman test at 95\% confidence level confirms that these differences are not significant. The two best performing user profiles ($<$$f0.9\_a12$$>$ and $<$$f0.8\_a12$$>$) only filter by fitness and eliminate up to 1/3 of the algorithms (corresponding to 12 algorithms). In contrast, two profiles with an aggressive time threshold strategy ($<$$f0.9\_t0.5\_a12$$>$ and $<$$f0.8\_t0.5\_a12$$>$), which also eliminate 1/3 of the algorithms, provide the lowest fitness at the end of the search. In general, we find that, regardless of the fitness threshold, profiles using a lower time threshold tend to perform the worst.

\begin{figure}
    \centering
    \begin{subfigure}[b]{1\textwidth}
         \centering
         \includegraphics[width=\textwidth]{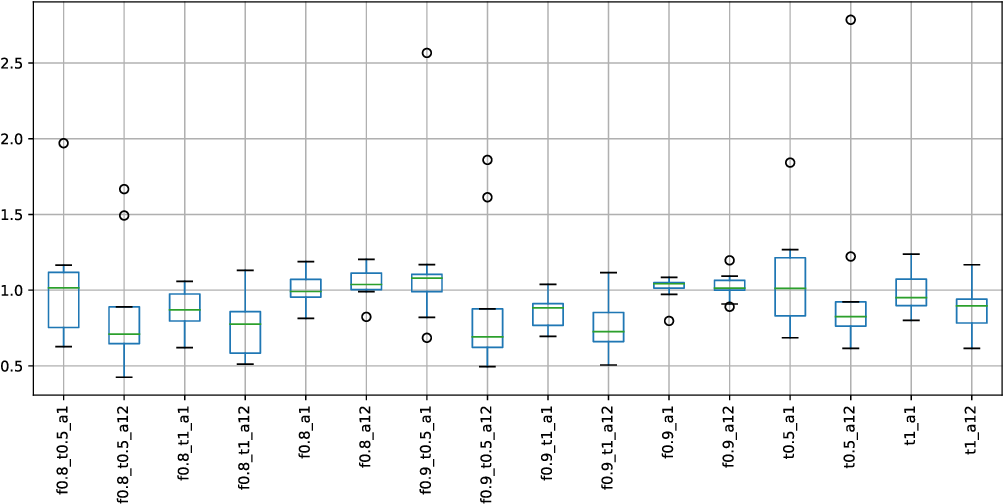}
         \caption{Speedup achieved by each simulated profile.}
         \label{fig:speedup-profile}
    \end{subfigure}
    \\
    \begin{subfigure}[b]{0.7\textwidth}
         \centering
         \includegraphics[width=\textwidth]{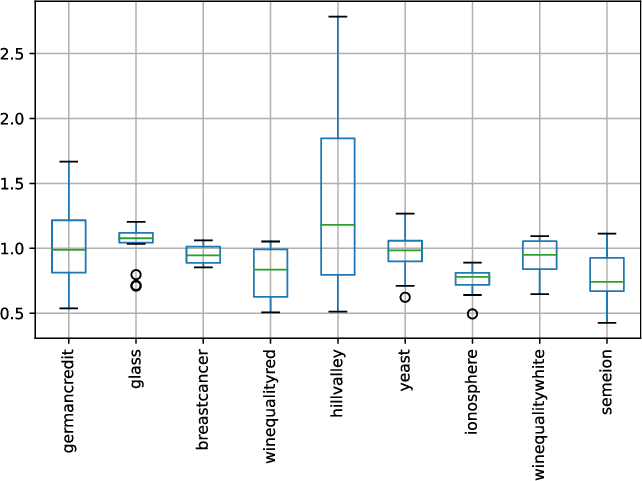}
         \caption{Speedup achieved per dataset.}
         \label{fig:speedup-dataset}
    \end{subfigure}
    \caption{Speedup achieved with simulated user interactions.}
    \label{fig:speedup}
\end{figure}

As the interaction is also intended to reduce workflow evaluation time, we analyse the speedup achieved due to the simulated users. Figures~\ref{fig:speedup-profile} and~\ref{fig:speedup-dataset} show boxplots with the distribution of the speedup achieved by user profile and dataset, respectively. The speedup is computed for the 30 repetitions based on the last 35 generations, as the first 15 generations are not affected by interactions. Focusing on the user profiles (Figure~\ref{fig:speedup-profile}), $<$$f0.9\_t0.5\_a12$$>$ and $<$$f0.8\_t0.5\_a12$$>$ achieve the largest speedups, requiring 69.1496\% and 70.9903\% of the time spent by the baseline, respectively. In fact, $<$$f0.9\_t0.5\_a12$$>$ only requires 42.4759\% of the time taken by the baseline in the \textit{semeion} dataset, which is the highest speedup among all simulated profiles and datasets. The \textit{semeion} dataset has the highest number of features and the second highest number of instances among all selected datasets. For this dataset, the baseline tends to include \textit{mlpClassifier} in the best workflows, which has a high evaluation time. In contrast, the $<$$f0.9\_t0.5_a12$$>$ strategy eliminates this classifier in 23 out of 30 runs, thus focusing on workflows with faster classifiers after the interactions.

We also observe that six profiles lead to an increase in the median evaluation time when compared to the baseline (median above 1 in the boxplot). Among them, we find profiles that apply the most aggressive time thresholds and only discard one algorithm ($<$$f0.9\_t0.5\_a1$$>$, $<$$f0.8\_t0.5\_a1$$>$ and $<$$t0.5\_a1$$>$) or only decide based on fitness ($<$$f0.9\_a1$$>$, $<$$f0.8\_a12$$>$ and $<$$f0.9\_a12$$>$). The first case suggests that eliminating the most costly algorithms does not guarantee a speedup at the end of the search. To illustrate this case, we focus on the \emph{hillvalley} dataset. This dataset exhibits very different results, as shown in Figure~\ref{fig:speedup-dataset}. For this dataset, our baseline method usually obtains the best results with the \emph{kNN} algorithm as classifier. However, simulated users tend to eliminate this classifier in the early stages due to its low predictive performance and its cost, which, although not high, is not among the best. As a consequence, \ourmethod loses the opportunity to combine it with the preprocessing algorithms that makes the workflow achieve good predictive performance. Furthermore, \ourmethod switches to selecting \emph{mlpClassifier}, which is even more expensive than \emph{kNN}. The \emph{mlpClassifier} classifier works well in combination with \emph{fastICA} for the \emph{hillvalley} dataset, which increases the dimensionality of the dataset and can therefore contribute to even longer evaluation times.

To summarise the results of this experiment, we conclude that \ourmethod can be guided to different solutions depending on the interaction strategies applied with the different user profiles. The interactions do not greatly affect the accuracy of the workflows, which suggests that \ourmethod is able to find alternative workflows with similar performance, even if the algorithms initially considered in good solutions are eliminated. More varied behaviour is observed with respect to evaluation time. In this case, the fact that the user profiles simulate simple heuristics, which do not change over time, implies that rigid decisions (\eg always eliminating algorithms) may not be a good strategy. This underlines the need for experiments with humans, who are expected to apply more complex heuristics to decide which algorithms to remove and how.

\subsection{Experiment 2: Human interaction}
\label{subsec:results-exp2}

In this section, we analyse the results obtained from the user study to answer RQ2-RQ4. First, we provide an overview of how the interactive sessions went in terms of interactions and generations used by the different participants. Data collected from the application logging component is used for this analysis. Participants chose to execute between 3 (one participant) and 9 interactions (three participants) out of the total number of available interactions (10). The majority of participants (80\%) performed more than 5 interactions, with the most frequent range of values being between 7 and 9 interactions (50\%). Regarding the distribution of the number of generations, 65\% of the participants performed all 35 available generations. Only 15\% of the participants stopped the process early, after evolving less than 25 generations. From these statistics, we confirm that the interactive experiment was successfully completed by all participants, and different user behaviours can be observed.

Another general aspect of the interactive experiment concerns the average time spent by participants in the interactions. 
Participants took significantly longer to complete the first and second interactions. In particular, one participant took more than 15 minutes to make a decision. This is probably due to unfamiliarity with the graphical environment and a larger decision space as to which algorithms and hyperparameters could be discarded. The time remains fairly constant afterwards, specially after the fifth interaction. This suggest that fatigue does not seem to be a pressing factor during the experiment.


\subsubsection{RQ2: Removal of algorithms and hyperparameter values}\label{subsec:exp2-rq2}

To answer RQ2, we inspected which algorithms and hyperparameter values were deleted most frequently and when such decisions were made. As for the most frequently deleted algorithms, we found that the following algorithms were deleted by at least 75\% of the participants: \textit{multinomialNB} (20 times), \textit{lda} (20), \textit{mlpClassifier} (19), \textit{decisionTree} (17), \textit{truncatedSVD} (17), \textit{logisticRegression} (16), \textit{passiveAggressiveClassifier} (16) and \textit{fastICA} (15). We highlight the high degree of agreement among users on which algorithms should be eliminated, most of them being classification algorithms. Despite the fact that the grammar includes a greater number of pre-processing options (20) than classifiers (15), pre-processing algorithms were eliminated 160 times, while classifiers were chosen for elimination 189 times. Probably, users perceive that the classifiers have more influence on the predictive capability of the workflows. On the other hand, it is worth noting that all algorithms have been removed at least once. This implies that the catalogue of algorithms is broad enough to accommodate the different preferences of the participants.

Focusing on the hyperparameter values, at least half of the users have removed the following: \textit{ward} (15) from \textit{fagg$::$linkage}, \textit{minority} (13) from \textit{rus$:: $sampling$\_$strategy}, \textit{l1} (13) of \textit{lr$::$penalty} and \textit{l1} (10) of \textit{lsvc$::$penalty}. In this case, we observe not only less agreement among users, but also less interest in modifying the grammar in terms of hyperparameter values. In fact, 48\% of the hyperparameter values have been removed by only one or two users.

\begin{figure}
    \centering
    \begin{subfigure}[b]{0.95\textwidth}
         \centering
         \includegraphics[width=\textwidth]{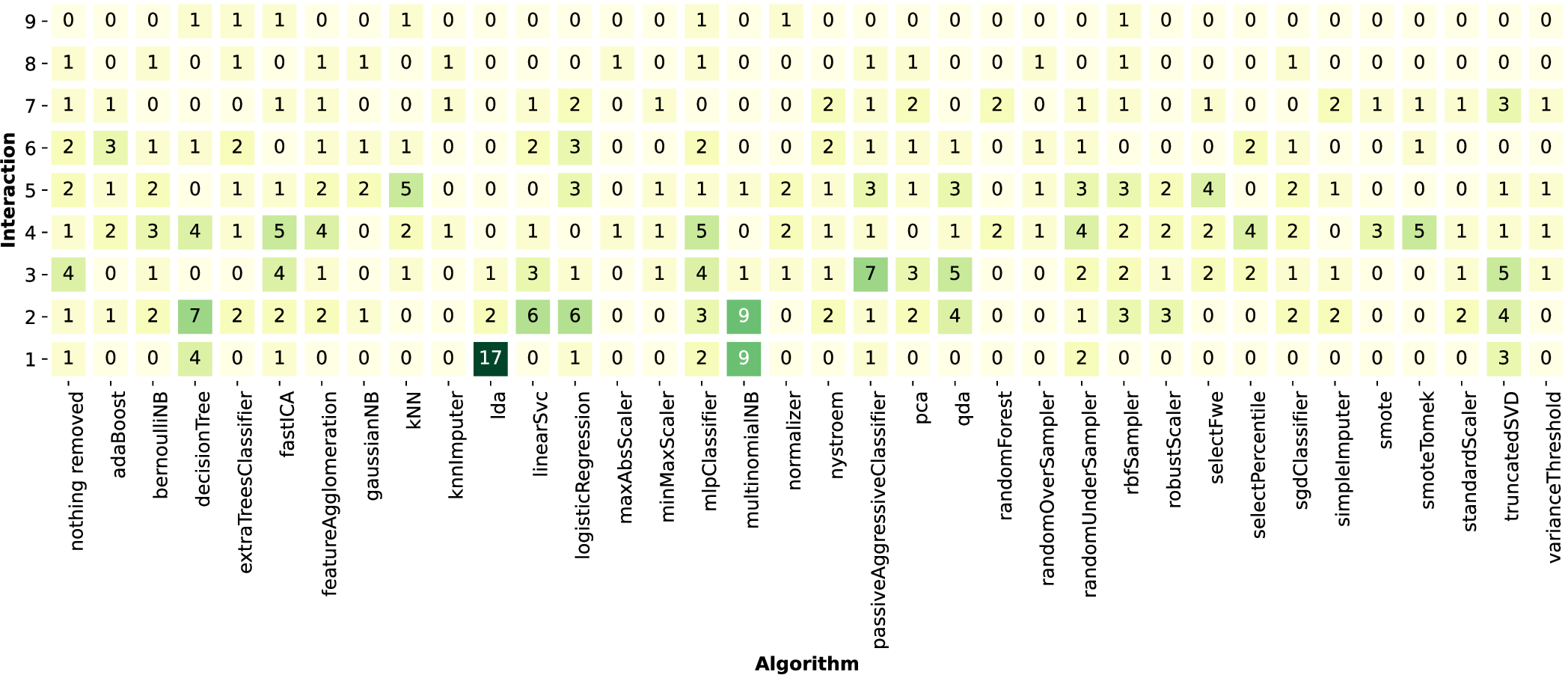}
         \caption{Classification and preprocessing algorithms.}
         \label{fig:del-algs}
    \end{subfigure}
    \hfill
    \begin{subfigure}[b]{0.95\textwidth}
         \centering
         \includegraphics[width=\textwidth]{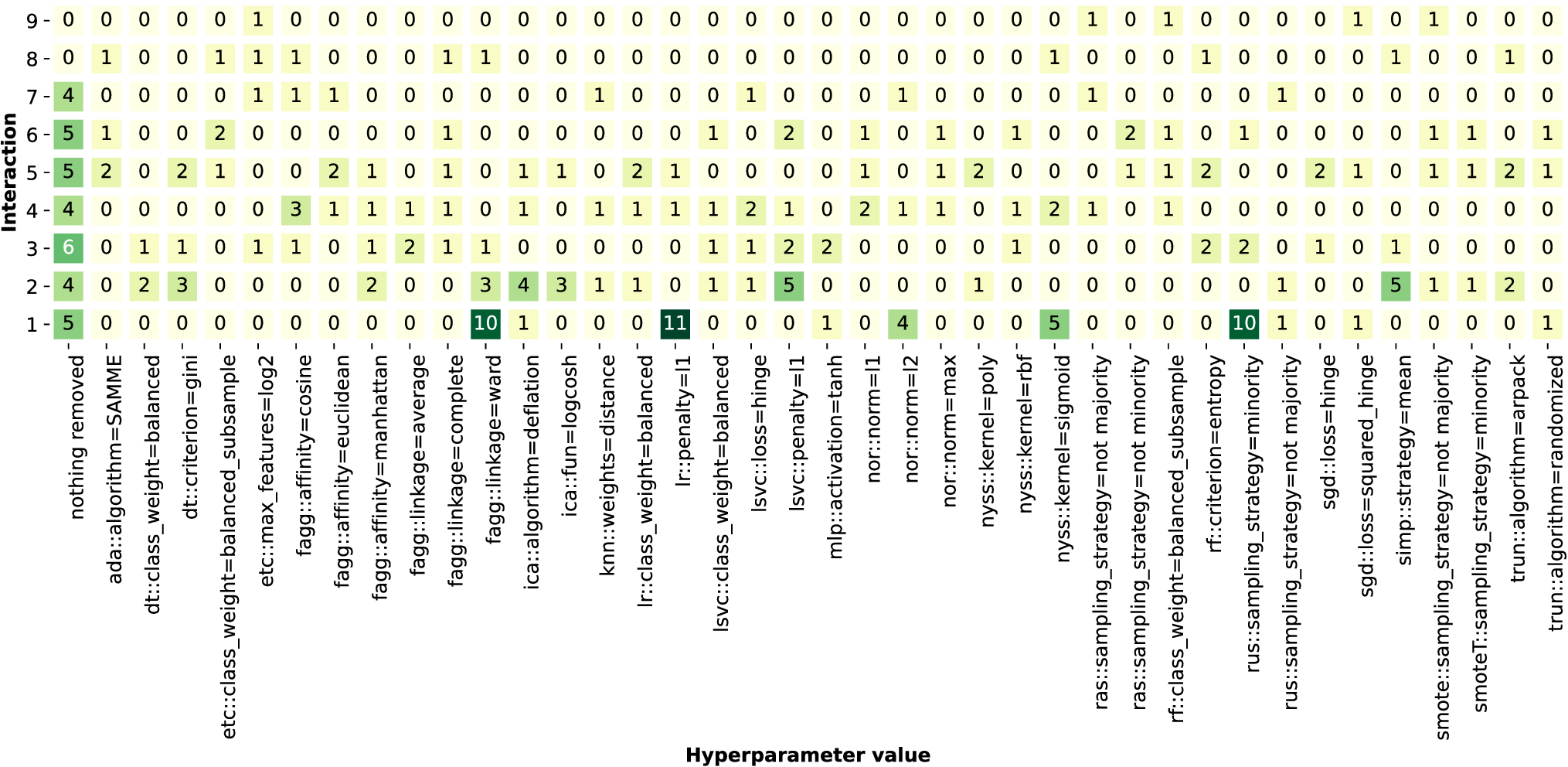}
         \caption{Hyperparameter values.}
         \label{fig:del-hps}
    \end{subfigure}
    \caption{Number of times each algorithm or hyperparameter value has been deleted by participants in each interaction.}
    \label{fig:rq2-del}
\end{figure}

In terms of how user actions are distributed, Figures~\ref{fig:del-algs} and~\ref{fig:del-hps} show how many times each algorithm and hyperparameter value has been deleted in each interaction (from 1 to 9), respectively. For reasons of space and readability, Figure~\ref{fig:del-hps} shows only those hyperparameter values that have been deleted by at least three participants. Focusing on the classification algorithms (Figure~\ref{fig:del-algs}), we observe that, in general, the most frequently deleted classifiers tend to be discarded in the first interactions. This is the case for two classifiers deleted by all participants (\textit{lda} and \textit{multinomialNB}): 90\% of the eliminations were in the first two interactions. The reason seems to lie in the fitness values achieved by the workflow containing both classifiers. On the one hand, workflows using \textit{lda} do not provide valid solutions (fitness equals 0 in these cases). On the other hand, the maximum fitness value reached by a workflow using \textit{multinomialNB} is 0.71, far away from the fitness value (0.94) of the best solution found by \ourmethod before pausing for the first interaction.

Other algorithms eliminated by the majority of participants (75\% or more) are also frequently discarded in the first interactions, although they also become the user's choice later. Compared to \textit{multinomialNB}, workflows in which these algorithms appear obtain better fitness values: \textit{decisionTree} (0.78), \textit{truncatedSVD} (0.79), \textit{passiveAggressiveClassifier} (0.81), \textit{logisticRegression} (0.83), \textit{mlpClassifier} (0.90) and \textit{fastICA} (0.94). Therefore, it seems reasonable to think that participants preferred to see if these algorithms could help build workflows closer to or better than the optimum found by \ourmethod prior to their intervention. In fact, \textit{mlpClassifier} provides similar performance and \textit{fastICA} is part of the best workflow found in the first 15 generations. The high removal rate in these cases is also related to their evaluation time. The average evaluation time of the workflows found before the first interaction is 0.85 seconds, but the best workflow ---which includes \textit{fastICA}--- requires 16.24 seconds. Most of the participants adopt a conservative strategy and decide to eliminate this algorithm after the fourth interaction, and some of them even keep it until the last interaction. Therefore, we can conclude that users prefer to eliminate algorithms guided by the predictive performance of the workflows containing them at the beginning, and evaluation times become more relevant at the end of the process. Apart from \textit{fastICA}, \textit{randomUnderSampler} and \textit{truncatedSVD} were the only preprocessing algorithms deleted in the first interaction. The choice of \textit{randomUnderSampler} is probably due to the fact that this algorithm frequently appears in the workflows with higher evaluation time but does not have a top predictive performance (0.86). In contrast, we suspect that \textit{truncatedSVD} is early discarded because of the solutions using it have a fitness lower than 0.79 as the execution time is below the average. In addition, Figure~\ref{fig:del-algs} indicates us that the removals of preprocessing algorithms are more scattered among the interactions. They probably consider that a low fitness is mainly due to the classifier, which makes it more difficult to decide which preprocessing algorithm should be removed.

Finally, looking at Figure~\ref{fig:del-hps}, we observe less agreement among users as to which hyperparameter values should be discarded and when. Here, the hyperparameter value discarded by the largest number of participants (\textit{ward} of \textit{fagg::linkage}) was chosen by 75\% of them. We also found interactions in which none of the participants eliminated any hyperparameter value, especially in the first six interactions. This type of decision changed at the end of the process, as at least one hyperparameter value was always removed in the last two interactions. We hypothesise that users who perform more interactions become more disruptive with respect to hyperparameter values because they have few interactions left. In other situations, users decided to remove the entire algorithm after consecutively removing some of their hyperparameter values. Such an strategy was observed in nine users after their fourth interaction. For example, most of the deletions of the \textit{logisticRegression} algorithm occurred in the second interaction, but one of its hyperparameter values (\textit{l1} for \textit{lr$::$penalty}) was frequently deleted (84\% of the time) in the first interaction. Another similar case occurs with the \textit{minority} value of the sampling strategy of a balancing algorithm (\textit{rus$::$sampling$\_$strategy}).

\subsubsection{RQ3: Impact of human intervention}\label{subsec:exp2-rq3}

To answer RQ3, we contextualise the human results with respect to those obtained by \ourmethod without interaction, which will be referred to as \emph{baseline}. In this way, we can analyse how interactions influence both the accuracy and the evaluation time that \ourmethod would have achieved if no grammar symbols were removed. The best workflow found by the baseline at the time interactions start (15 generations) has a fitness value of 0.9365 and its evaluation requires 16.25 seconds. After 50 generations, the best workflow has a balanced accuracy equal to 0.9406 and its evaluation time is 16.08 seconds. In total, the baseline took 53.39 minutes to complete all 50 generations.

We have analysed the trade-off between fitness and evaluation time achieved by participants, and whether their best workflow improves on either criterion over that found by the baseline. Four situations can occur: (a) both fitness and evaluation time improve, (b) only fitness improves, (c) only evaluation time improves and, (d) neither fitness nor evaluation time improves. Focusing on case (a), seven participants help \ourmethod to find a workflow with better fitness and shorter evaluation time than the baseline. The average fitness is 0.9455, while the average evaluation time is 0.84 seconds. This represents a 19.19 of speedup in terms of workflow evaluation time. The best workflow found by one participant has a fitness of 0.9598 and an evaluation time of 0.12 seconds, representing a speedup of 129.90 with respect to the baseline. This participant eliminated \textit{fastICA} and \textit{extraTreeClassifier}, two algorithms included in the best workflow found by \ourmethod in the first 15 generations. The participant discarded both algorithms once he found a better workflow that included less computationally expensive algorithms. These algorithms were also eliminated by the other six participants ---\textit{fastICA} by four of them and \textit{extraTreeClassifier} by one of them. However, this other elimination of \textit{extraTreeClassifier} occurred in the last interaction, after which \ourmethod only ran two more generations and probably did not have the same effect on the search.

One participant is in case (b), which means that the fitness improves at the expense of a longer evaluation time. The fitness value (0.9498) corresponds to the second best workflow found by any participant and the evaluation time (16.52) is almost the same as the solution found by the baseline (16.25). This participant followed a rather conservative strategy, focusing on exploiting the region of the search space around the best solution found by \ourmethod. The discarded algorithms and hyperparameter values had low predictive ability, and those that appeared in the best solution shown in the first interaction remained in the grammar. In this case, the participant was not as concerned about evaluation time, as computationally expensive algorithms such as \textit{fatICA} remained throughout the search.

Focusing on case (c), eight participants guided the algorithm only to solutions with a shorter evaluation time. The average evaluation time is 0.6661 seconds, which represents a speedup of 24.14 over the baseline. This improvement comes at the expense of the fitness value, whose average value is 0.9282. This is a slight decrease (1.32\%) compared to the baseline solution. However, \ourmethod always saves the best solution found, so participants could choose between the global best solution ---the one they were shown in generation 15--- or the local optimum they found at the end of the search. Finally, four participants did not improve either fitness or evaluation time (case (d)). Therefore, the best workflow they have obtained is the same as the one they had before the first interaction. Three of the four participants acknowledged having little ML experience (only two years), so they may have less confidence in the impact of different algorithms and hyperparameter values on the workflow composition problem.

As a final observation, the average execution time during the experiment was 31.13 minutes, which represents an acceleration of 1.72 from the baseline. This value excludes the time taken by the participants to perform the interactions, which was 20.30 minutes on average. This makes the total time very similar to the baseline (53.39 minutes). This comparison should be analysed with caution, as not all participants completed all 50 generations and the reduced search space means that more repeated solutions may appear that do not require evaluation. In addition, the dataset selected for the interactive experiment was not the most expensive in order to reduce user fatigue. The time saved by the method can be expected to increase as the complexity of the dataset increases, while the time spent by a user could remain more or less constant.

\subsubsection{RQ4: User experience}\label{subsec:exp2-rq4}

\begin{figure}
     \centering
     \begin{subfigure}[b]{0.48\textwidth}
         \centering
         \includegraphics[width=\textwidth]{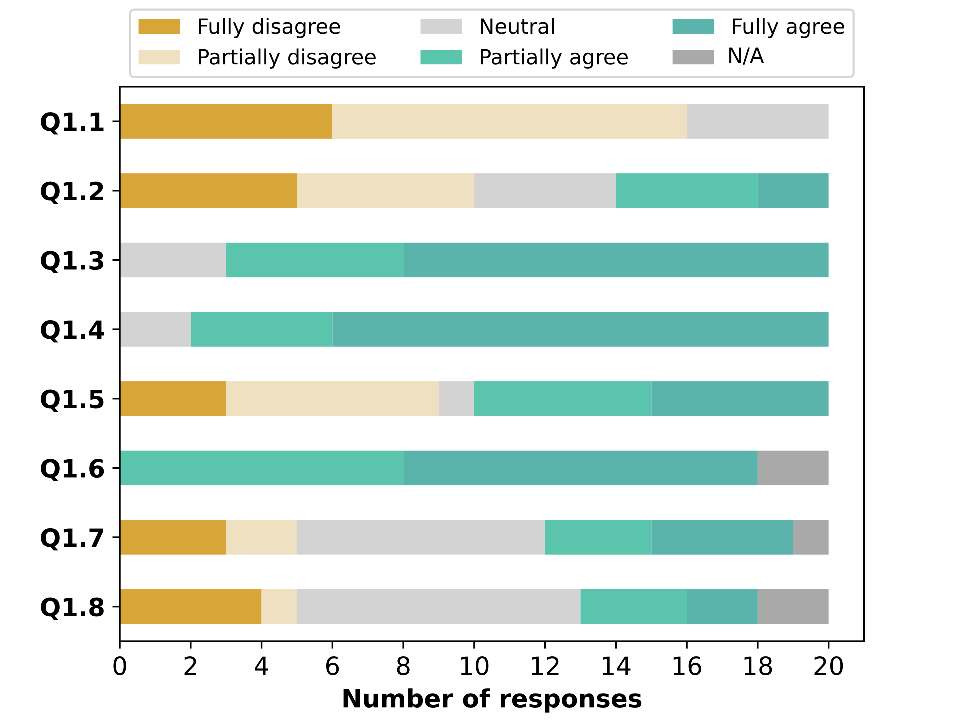}
         \caption{Optimisation process (Q1.1-Q1.8)}
         \label{fig:rq4-q1}
     \end{subfigure}
     \hfill
     \begin{subfigure}[b]{0.48\textwidth}
         \centering
         \includegraphics[width=\textwidth]{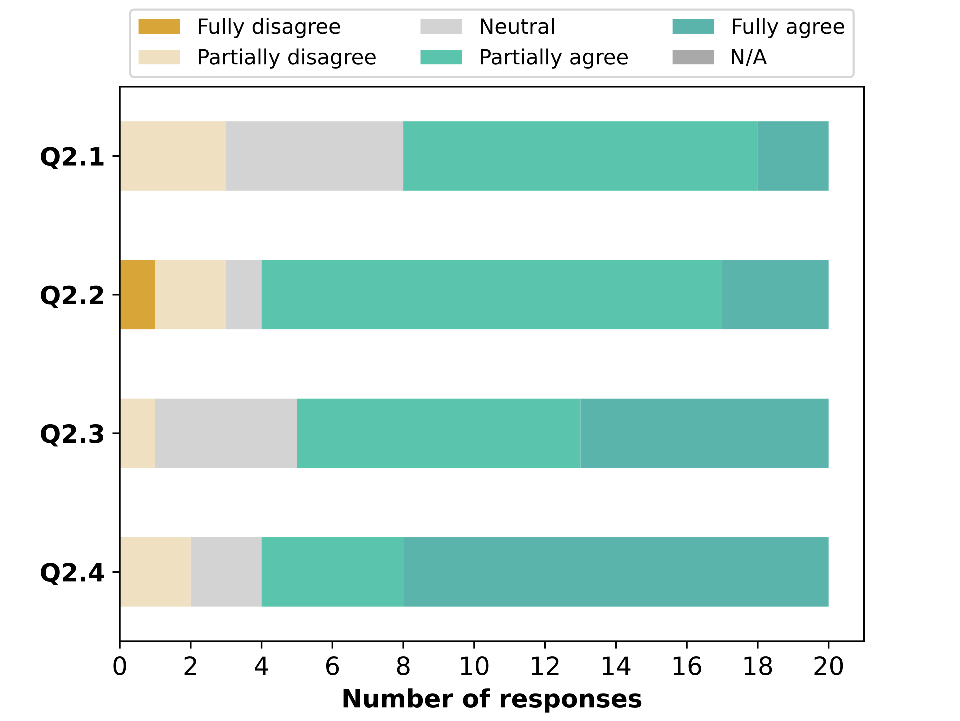}
         \caption{Evolution of the process (Q2.1-Q2.4)}
         \label{fig:rq4-q2}
     \end{subfigure}
     \hfill
     \begin{subfigure}[b]{0.48\textwidth}
         \centering
         \includegraphics[width=\textwidth]{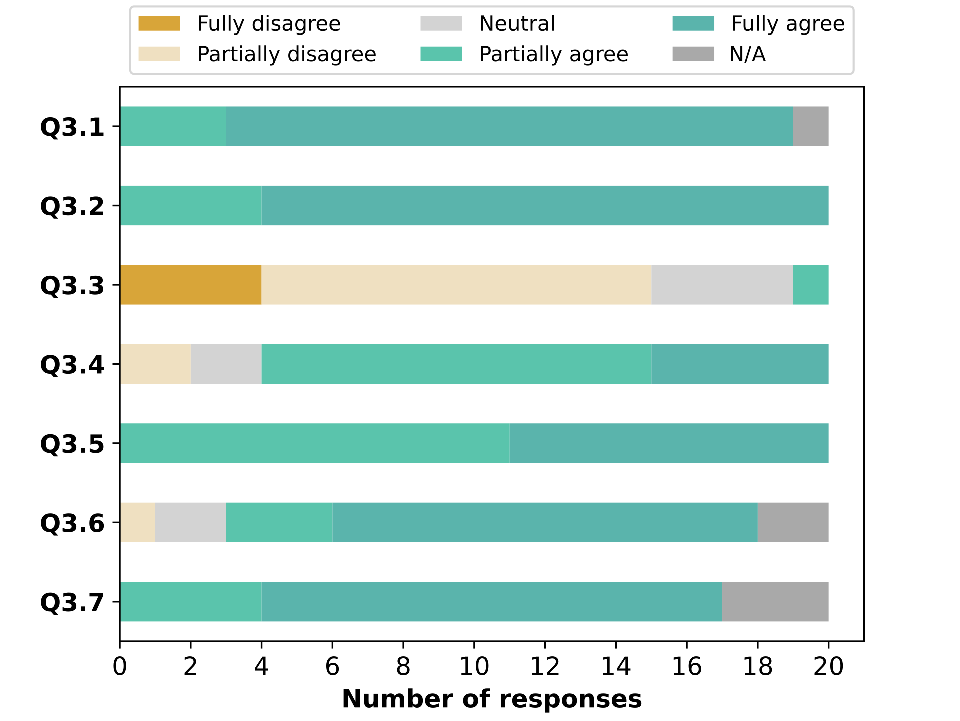}
         \caption{User experience (Q3.1-Q3.7)}
         \label{fig:rq4-q3}
     \end{subfigure}
    \begin{subfigure}[b]{0.48\textwidth}
         \centering
         \includegraphics[width=\textwidth]{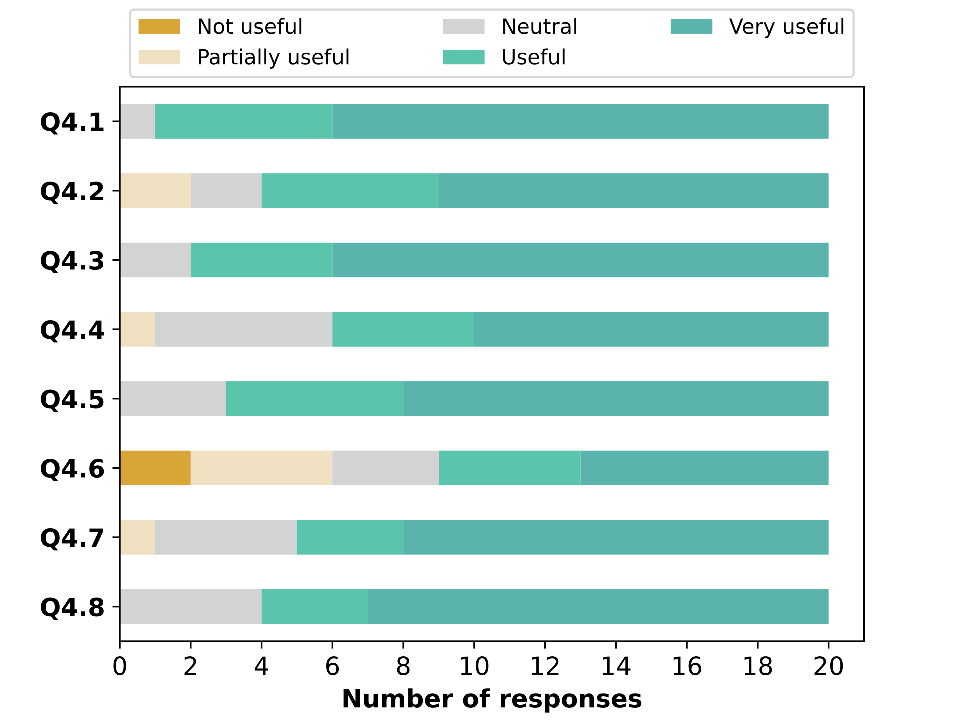}
         \caption{Usefulness (Q4.1-Q4.8)}
         \label{fig:rq4-q4}
     \end{subfigure}
     \begin{subfigure}[b]{0.48\textwidth}
         \centering
         \includegraphics[width=\textwidth]{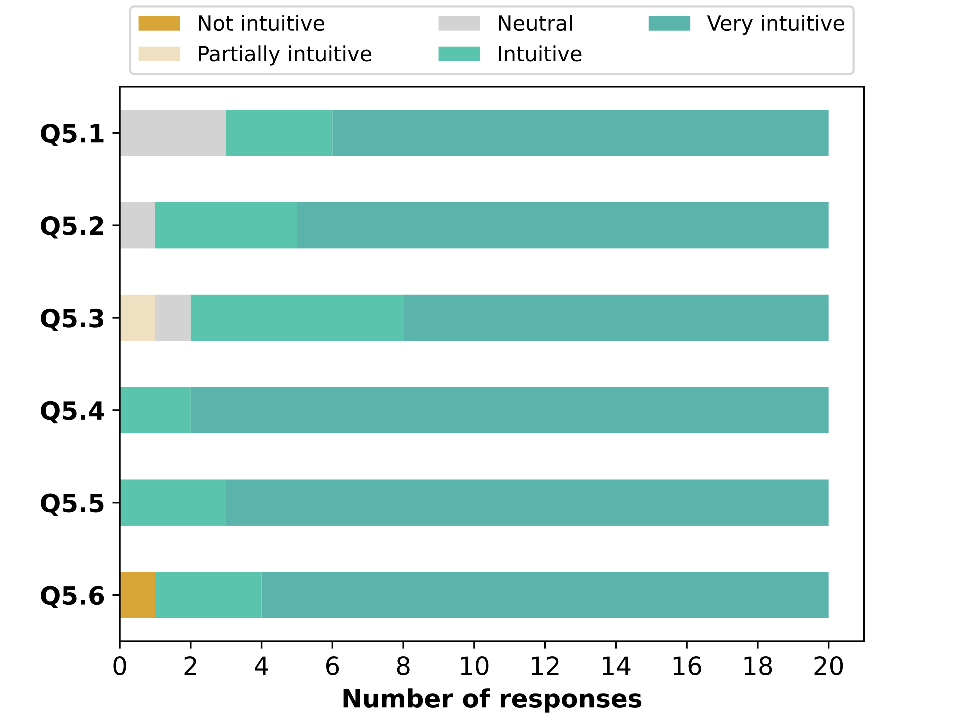}
         \caption{Intuitiveness (Q5.1-Q5.6)}
         \label{fig:rq4-q5}
     \end{subfigure}
        \caption{Responses of the participants to the questionnaire.}
        \label{fig:rq4-survey}       
\end{figure}

To analyse the user experience (RQ4), we present the answers to the questionnaire that the participants filled in after the experiment. Figure~\ref{fig:rq4-survey} shows the responses to the statements and questions listed in Table~\ref{tab:questions}.

\paragraph{Optimisation process} According to Figure~\ref{fig:rq4-q1}, none of the participants found the interactions tedious (Q1.1). Also related to fatigue, six participants (30\%) acknowledged that they paid more attention at the beginning of the interactions (Q1.2). This behaviour is common in interactive experiments and allows us to confirm that most participants remain interested. As for the graphical decision-support elements, the scatter plot showing the trade-off between accuracy and evaluation time was useful for 85\% of the participants (Q1.3). 90\% of the participants agree that the table with detailed statistics was very effective in helping them to eliminate grammar symbols (Q1.4). This gives us confidence that the graphical interface was fit for the purpose of the interaction.

In contrast, the time plot comparing total evaluation time against baseline was considered less relevant for decision making (Q1.5), with only half of the participants partially or fully agreeing with this aspect. Possible causes are that participants were more focused on fitness improvement and that the time information shown in the table was sufficient to choose which algorithms and hyperparameter values to eliminate.

In fact, none of the participants missed additional information that would help in decision-making (Q1.6). One participant indicated in the free comments that the scatter plot could be improved by shading the algorithms and values that are proposed to the user for deletion. The last two questions refer to the participant's preferences in terms of the type of grammar element to be removed. There is no clear trend in either Q1.7 ---preference for removing algorithms instead of hyperparameter values--- or Q1.8 ---preference for removing classifiers instead of preprocessing algorithms. In this sense, we can conclude that our set of participants was diverse in the strategies they followed to interact, and that having the possibility to remove different elements from the workflow was adequate to gain flexibility.

\paragraph{Evolution of the process} Focusing on Figure~\ref{fig:rq4-q2}, the majority of participants (60\%) commented that the algorithm worked as expected (Q2.1). Among the three participants who partially disagreed with this statement, one of them did not achieve a workflow with better fitness or evaluation time than the solution initially shown. The other two participants achieved some improvement after the first interaction, but the algorithm did not locate a better workflow after that. Therefore, a sense of stagnation could be the cause of their response. Regarding Q2.2, 80\% of the participants agree that the algorithm showed solutions they did not expect. Again, those who disagree with this statement are participants whose interaction did not allow the algorithm to move away from the initial solution shown or only achieved a small improvement. Overall, we noted satisfaction with the performance of \ourmethod, as it helped participants to explore novel workflows.

Questions Q2.3 and Q2.4 refer to how participants perceive the impact of their actions. 75\% of the participants partially or fully agree with the idea that interactions have an effect on the composition of new workflows (Q2.3). A higher percentage of participants (80\%) gave positive responses to Q2.4, which focused on the impact of their action on workflow evaluation time. The log files of participants who disagreed to some extent with these questions reveal that they made small improvements in both fitness and evaluation time compared to other participants, and that such improvements were mostly focused on the first interaction. Again, the fact that participants were more or less conservative in terms of which algorithms and hyperparameter values to eliminate may change their perception of the search process.

\paragraph{User experience} The answers to questions Q3.1-Q3.7 are summarised in Figure~\ref{fig:rq4-q3}. The vast majority of participants (95\%) find interactive assistants for workflow composition interesting (Q3.1). Only one participant, an undergraduate student with one year of ML experience, did not answer this question. All participants agree that the information displayed by the tool was useful (Q3.2) and only one participant missed some additional information to support decision making (Q3.3). We found some free comments in this sense from other participants, suggesting to include the use of RAM or other resources required by each algorithm.

80\% of the participants responded positively to question Q3.4, meaning that they were satisfied with the time spent on the process in view of the results obtained. Those who show some disagreement are the participants who did not improve the initial solution with their decisions. As already mentioned, the user's perception of the process is strongly influenced by the feeling of progress in the search and the exploration of new regions of the search space. Despite this, all participants acknowledge that \ourmethod has allowed them to discover new ways of composing workflows (Q3.5). There were also very positive responses to Q3.6 ---interest in using \ourmethod with another dataset--- and Q3.7 ---interest in using \ourmethod to solve their own problems. Participants with neutral or no responses to these questions have little ML experience. We hypothesise that they might consider workflow composition as a complex problem for which they need more knowledge to get the most out of the tool. 

\paragraph{Usefulness} Figure~\ref{fig:rq4-q4} summarises the responses related to usefulness (Q4.1-Q4.8). Overall, the majority of the participants agree that all the elements included in the interface to support decision making were useful. The interactive scatter plot (Q4.1) is the most positively valued by the participants (95\%). Although there is little difference, participants have focused more on the table statistics related to accuracy (Q4.3 and Q4.5) than those related to evaluation time (Q4.2 and Q4.4). The line plot showing the total time spent with and without interaction is the element that participants found less useful (55\%). The reason may be that it shows information on what has already been done, \ie the time gained due to grammar reduction so far. In fact, participants were interested in the reduction of evaluation time according to answers Q4.2 and Q4.4: 80\% and 85\% of participants found it useful to know the average evaluation time of workflows that include a specific algorithm or hyperparameter value. Finally, at least 75\% of the participants appreciated the possibility of indicating the number of generations after each interaction, as well as being able to end the execution after each interaction.

\paragraph{Intuitiveness} Figure~\ref{fig:rq4-q5} confirms that all elements of the interactive tool are very intuitive (P5.1-Q5.6). The options for deleting algorithms (Q5.4) and hyperparameter values (Q5.5) are clearly aligned with the purpose of the experiment, so all participants rated them as intuitive or very intuitive. Most participants also rated positively the options for distributing the number of generations (Q5.1, 85\%) and manipulating the scatter plot (Q5.2, 95\%). One participant did not consider the use of thresholds as intuitive (Q5.3), probably because it uses relative values, as also mentioned by other participants. Only one participant indicated that stopping the search was not an intuitive action (Q5.6). 
Also related to the use of the tool, some participants propose some improvements to enhance the user experience: 1) adding a context menu that includes information on algorithms and hyperparameter values;\footnote{In the interactive experiment, this information was provided in a document.} and 2) highlighting the algorithms and hyperparameter values that \ourmethod recommends to be removed depending on the thresholds set by the participant.

\section{Concluding remarks}
\label{sec:conclusions}

This paper presents \ourmethod, an interactive grammar-guide genetic programming algorithm designed to address automatic workflow composition under a ``human-in-the-loop'' approach. Human intervention seeks to refine the grammar so that the search space is pruned according to human preferences. \ourmethod is able to redirect the search by excluding workflows that include algorithms and hyperparameter values that do not meet the user's expectations in terms of accuracy, evaluation time or both. Thus, \ourmethod can be applied by a wide variety of user profiles, from those who show prevalence to certain algorithms to users who prefer to sacrifice accuracy in favour of tight time constraints.

Our experiments show that \ourmethod responds quickly to modifications of the search space caused by the adaptation of the grammar. Under different simulated user profiles, we evaluate how the trade-off between predictive accuracy and evaluation time can be controlled, showing an improvement of accuracy of up to 0.7166\% 
and an speedup of up to 2.3542 compared to a non-interactive run, respectively. More interestingly, we conducted a real interactive experiment with 20 participants to empirically validate the benefits of human intervention in automatic workflow composition. Seven participants were able to guide the algorithm to unexplored solutions with higher accuracy (up to 2.043\% of improvement) and shorter evaluation time (maximum speedup of 5.7015) than the baseline, while nine participants improved at least one of these two aspects. All agreed that the interaction with the method allowed them to gain knowledge about the AWC problem. Thus, interactive methods such as ours may attract more attention to AutoML, as users perceive that their preferences can positively influence the solutions obtained.

In the future, we plan to increase the interactive capabilities of \ourmethod to support more user actions. These actions could include manual editing of parts of the workflow or the possibility to freeze some of its elements. The graphical interface could also be improved to include suggestions made by participants in the free comments, such as reintroducing deleted algorithms and including complementary information about the algorithms.

\section*{Additional material}
The definition of the grammar and the full experimental results are publicly available. For the user study, we provide the questionnaire, the anonymised answers and the log files generated by the participants. Screenshots of the interactive tool are also included. All this complementary material can be accessed from the following Zenodo repository: \textit{anonymized while under review} 

\section*{Acknowledgements}
The authors would like to thank the participants for their
valuable time, strong interest and positive feedback. This work was supported by the MICIN/AEI/10.13039/501100011033 under Grant PID2020-115832GB-I00; and Junta de Andaluc\'ia (postdoctoral contract DOC\_00944).

\bibliographystyle{elsarticle-num} 
\bibliography{references}

\end{document}